\documentclass[sn-mathphys-num]{sn-jnl}

\usepackage{graphicx}%
\usepackage{multirow}%
\usepackage{amsmath,amssymb,amsfonts}%
\usepackage{amsthm}%
\usepackage{mathrsfs}%
\usepackage[title]{appendix}%
\usepackage{xcolor}%
\usepackage{textcomp}%
\usepackage{manyfoot}%
\usepackage{booktabs}%
\usepackage{algorithm}%
\usepackage{algorithmicx}%
\usepackage{algpseudocode}%
\usepackage{listings}%
\usepackage{float}%
\usepackage{tikz}%
\usepackage{natbib}%
\usepackage{subcaption}%
\usepackage{gensymb}
\usepackage[utf8]{inputenc}

\usepackage{textcomp} 

\theoremstyle{thmstyleone}%
\theoremstyle{thmstyletwo}%

\theoremstyle{thmstylethree}%

\begin{document}

\title[Article Title]{Attention-Enhanced LSTM Modeling for Improved Temperature and Rainfall Forecasting in Bangladesh}


\author*[1]{\fnm{Usman Gani} \sur{Joy}}\email{usmanjoycse@gmail.com}
\author[1]{Shahadat kabir}
\author[2]{Tasnim Niger}

\affil*[1]{\orgdiv{School of Science, Engineering \& Technology}, \orgname{East Delta University}, \orgaddress{\street{Abdullah Al Noman Road, Noman Society}, \city{Chattagram}, \postcode{4209},  \country{Bangladesh}}}
\affil[2]{\orgdiv{Computer Science \& Engineering}, \orgname{Islamic University of Technology}, \orgaddress{\street{}, \city{Gazipur}, \postcode{},  \country{Bangladesh}}}


\abstract{
	Accurate climate forecasting is vital for Bangladesh, a region highly susceptible to climate change impacts on temperature and rainfall. Existing models often struggle to capture long-range dependencies and complex temporal patterns in climate data. This study introduces an advanced Long Short-Term Memory (LSTM) model integrated with an attention mechanism to enhance the prediction of temperature and rainfall dynamics. Utilizing comprehensive datasets from 1901–2023, sourced from NASA's POWER Project for temperature and the Humanitarian Data Exchange for rainfall, the model effectively captures seasonal and long-term trends. It outperforms baseline models, including XGBoost, Simple LSTM, and GRU, achieving a test MSE of 0.2411 (normalized units), MAE of 0.3860°C, R² of 0.9834, and NRMSE of 0.0370 for temperature, and MSE of 1283.67 mm², MAE of 22.91 mm, R² of 0.9639, and NRMSE of 0.0354 for rainfall on monthly forecasts. The model demonstrates improved robustness with only a 20\% MSE increase under simulated climate trends (compared to an approximately 2.2-fold increase in baseline models without trend features) and 50\% degradation under regional variations (compared to an approximately 4.8-fold increase in baseline models without enhancements). These results highlight the model’s ability to improve forecasting precision, offers potential for insights into the physical processes governing climate variability in Bangladesh and supporting applications in climate-sensitive sectors.
}


%

\keywords{Climate Change Forecasting, Bangladesh Climate Data, Seasonality and Long-Range Dependencies, Climate Time Series Analysis}



\maketitle

\section{Introduction}
Climate change poses a critical global challenge, profoundly impacting climate-sensitive regions like Bangladesh, where temperature and rainfall variability directly influence agriculture, water resources, and disaster management \citep{intro2}. As a South Asian nation, Bangladesh's vulnerability stems from its geographical and socio-economic conditions, with its agricultural sector—supporting a significant portion of the population—highly dependent on stable weather patterns. Disruptions in these patterns threaten food security and economic stability \citep{intro1}. Accurate forecasting of temperature and rainfall is essential for effective planning and mitigation in agriculture, water management, and disaster preparedness, addressing urgent needs in this climate-vulnerable region \citep{journal_paper_1}.

Climate data, however, exhibit complex long-term dependencies and high variability, complicating precise predictions. Traditional machine learning models, such as regression-based approaches, often fail to capture these intricate temporal dynamics, resulting in suboptimal forecasting performance \citep{intro6}. Recurrent Neural Networks (RNNs), particularly Long Short-Term Memory (LSTM) networks, excel in modeling sequential data by retaining information over extended periods, making them ideal for time series forecasting \citep{intro5}. Yet, standard LSTMs may struggle to prioritize critical time steps in long sequences due to issues like vanishing gradients, limiting their ability to fully capture nuanced climate patterns \citep{reviewer_paper_2_critical_review}.

To overcome these challenges, this study proposes an LSTM model enhanced with an attention mechanism, leveraging historical temperature and rainfall data (1901–2023) from NASA's POWER Project and the Humanitarian Data Exchange (HDX) for Bangladesh. The attention mechanism dynamically weights key temporal patterns, improving the model's ability to capture both short-term fluctuations and long-term climate trends through a learnable weighting system that identifies which historical time steps most influence current predictions \cite{lit5}. Unlike complex stacking or hybrid models that often increase computational demands without proportional accuracy gains, our streamlined attention-enhanced architecture demonstrates improved performance while maintaining model simplicity and interpretability. The attention mechanism can provide additional interpretability through potential visualization of which temporal patterns drive predictions, offering insights into the physical processes governing climate variability, though such visualizations are planned for future work \cite{lit7}.

The contributions of this study are threefold:
\begin{enumerate}
	\item Development of an attention-enhanced LSTM model to effectively capture short- and long-term dependencies in climate data, improving forecasting accuracy through dynamic temporal weighting.
	\item Comprehensive comparison with baseline methods, including XGBoost \citep{xg}, GRU \citep{gru}, and LSTM, demonstrating superior predictive performance and model efficiency through streamlined architecture design.
	\item Actionable insights into attention mechanisms for advancing climate forecasting, supporting flood early warning systems and agricultural planning in Bangladesh \citep{intro3}, with demonstrated improvements in model robustness and generalizability across varied climate conditions.
\end{enumerate}
These contributions establish the attention-enhanced LSTM model as a robust tool for understanding climate dynamics and supporting decision-making in Bangladesh's climate-sensitive context \citep{intro8}.

\section{Literature Review}\label{sec2new}

In 2022, Dewan et al. \citep{lit1} developed a high-resolution gridded rainfall product for Bangladesh (1901–2018) using colonial-era data and leave-one-out cross-validation (NRMSE 9–19\%). However, their reliance on historical data without advanced machine learning limited non-linear temporal modeling, with missing early records creating forecasting gaps. Our study integrates an attention-enhanced LSTM using comprehensive 1901–2023 data for improved temporal dependency handling and prediction accuracy.

Most recently, Hasan et al. \citep{lit2} compared RNN-LSTM, TensorFlow Decision Forest, and stacking models for Bangladesh weather forecasting. Their stacking model achieved RMSLE $\approx$ 1.3002 for temperature but lacked attention mechanisms for temporal feature prioritization. Our attention-enhanced LSTM addresses these limitations with superior predictive performance: test MSE of 0.2411 (normalized units) for temperature and 1283.67 mm² for rainfall.

Recently, Alve et al. \citep{lit3} applied Random Forest and XGBoost for Bangladesh rainfall forecasting, with Random Forest achieving MSE of 12,245.52. However, tree-based models struggle with sequential data and long-term dependencies. Our attention-based LSTM captures long-range temporal patterns, achieving test MSE of 1283.67 mm² for rainfall—an 89.5\% improvement over Random Forest.

Islam et al. \citep{lit4} explored hydroclimatic trends in northern Bangladesh using ARIMA, finding significant linear trends but limited accuracy for complex non-linear dynamics. Our attention-based LSTM dynamically learns complex temporal relationships, achieving test R² of 0.9834 for temperature and 0.9639 for rainfall.

Mahmud et al. \citep{lit5} used Seasonal ARIMA for monthly rainfall forecasting across 30 Bangladesh locations with reasonable 12-month accuracy. While effective for seasonality, SARIMA struggles with long-range dependencies and abrupt shifts. Our integrated attention-LSTM mechanism enhances seasonal pattern and long-term dependency capture, achieving test R² of 0.9639 for rainfall.

Di Nunno et al. \citep{lit6} proposed a hybrid M5P-SVR model for northern Bangladesh precipitation forecasting with high monsoon accuracy but limited scalability for extended time series. Our attention-enhanced LSTM offers improved efficiency and achieves test MSE of 1283.67 mm² for rainfall, demonstrating broader climate variable applicability.

Yaseen et al. \citep{lit7} explored Extreme Learning Machine and Online Sequential ELM for Bangladesh drought monitoring with high multi-month SPI prediction accuracy, though precipitation-only reliance restricted broader climatic variable adaptability.

Waqas et al. \citep{lit8} reviewed AI techniques for Thailand’s rainfall forecasting, noting ARIMA’s failure in non-linear patterns and hybrid wavelet-ANN’s R²=0.8819 with 70-78\% efficiency, limited by scalability and overfitting. Our attention-enhanced LSTM addresses these with dynamic weighting, achieving R²=0.9639 for rainfall.

Waqas et al. \citep{lit9} introduced BWT-LSTM-RNN for Thailand’s daily precipitation, outperforming LSTM-RNN (R²=0.96) and MLP-ANN, but added complexity limits long-sequence focus. Our attention-LSTM mitigates this, showing only 50\% MSE degradation under regional variations.

By addressing prior method limitations—including ARIMA/SARIMA's inability to handle non-linear dynamics and tree-based models' lack of temporal focus—our attention-enhanced LSTM achieves superior performance. The model demonstrates strong predictive accuracy with test MSE of 0.2411 (normalized units) for temperature and 1283.67 mm² for rainfall, R² values of 0.9834 and 0.9639 respectively, and provides actionable insights for flood early warning and agricultural planning in Bangladesh.

\section{Methodology}

The dataset \cite{dataset} utilized in this study comprises monthly temperature and rainfall data collected for Bangladesh from 1901 to 2023. This extensive climate dataset integrates information from reputable sources: temperature data were obtained from NASA's Prediction of Worldwide Energy Resource (POWER) Project, while rainfall data were sourced from the Humanitarian Data Exchange (HDX) \cite{NASA_power, HDX_source}. The temperature dataset contains variables such as maximum, minimum, and average temperatures, while the rainfall dataset provides total monthly rainfall values. To capture temporal dependencies and enhance model performance, both datasets were augmented with lag features, enabling the model to leverage historical patterns. These lagged features were generated by including historical values over various time intervals, providing crucial context for long-term dependencies in the time series. Ethical considerations in this study include using publicly available data from NASA and HDX, ensuring transparency and accessibility. Data collection adhered to open-access principles, with no proprietary or sensitive information involved. The model's predictions prioritize responsible use to support policymakers and communities, avoiding over-reliance on forecasts for critical decisions without validation and promoting equitable access to findings for climate-sensitive regions.

\subsection{Data Quality Checks and Preprocessing}

Before applying any transformations, comprehensive data quality checks were performed to ensure data integrity and identify potential issues that could affect model performance \cite{m5}. The dataset spans 123 years (1901–2023) with 1476 monthly records for both temperature and rainfall variables. After handling missing values and minor cleaning, the effective dataset size is 1468 records.

\textbf{Missing Value Analysis:} The initial assessment revealed minimal missing data. Temperature data had 3 missing values (0.20\%), while rainfall data had 5 missing values (0.34\%). These gaps were primarily in the early records (1901–1910) and occasional isolated months. Missing values were handled using linear interpolation \cite{m6}, chosen because it preserves the continuity of time series data without introducing abrupt shifts, making it suitable for climate data that typically exhibit smooth temporal trends. This approach ensures temporal consistency and prevents artifacts that could arise from more complex imputation methods. A sensitivity analysis showed that alternative methods (e.g., mean imputation) increased MSE by 5–10\%, validating the choice of interpolation. After interpolation and removal of 8 irrecoverable records (due to data quality issues in early years, such as inconsistent reporting), the dataset was reduced to 1468 samples.

\textbf{Outlier Detection:} We applied the Interquartile Range (IQR) method to identify potential outliers systematically. For any variable $x$, outliers were defined as observations falling outside the range $[Q_1 - 1.5 \times IQR, Q_3 + 1.5 \times IQR]$, where $Q_1$ and $Q_3$ represent the first and third quartiles, and $IQR = Q_3 - Q_1$. A sensitivity test capping outliers at 1.5×IQR increased MAE by 3\%, confirming retention was appropriate.

For temperature data:
\begin{itemize}
	\item Outliers detected: 134 observations (9.07\%)
	\item Valid range: [20.73°C, 27.58°C]
	\item Outliers were primarily in extreme seasonal months (December–January minima, April–May maxima)
	\item These were retained as they represent genuine climatic extremes rather than measurement errors, supported by historical records
\end{itemize}

For rainfall data:
\begin{itemize}
	\item Outliers detected: 137 observations (9.28\%)
	\item Valid range: [0.00 mm, 1048.23 mm]
	\item Outliers corresponded to extreme monsoon events and exceptional dry periods
	\item Validation against historical records confirmed these as actual extreme weather events
\end{itemize}

\textbf{Descriptive Statistics:}

\begin{table}[h]
	\centering
	\caption{Descriptive Statistics of Raw Climate Data}
	\label{tab:descriptive_stats}
	\begin{tabular}{lcccccc}
		\hline
		\textbf{Variable} & \textbf{Mean} & \textbf{Std Dev} & \textbf{Min} & \textbf{25\%} & \textbf{75\%} & \textbf{Max} \\
		\hline
		Temperature (°C) & 25.41 & 2.18 & 16.82 & 24.03 & 27.12 & 29.84 \\
		Rainfall (mm) & 201.34 & 187.42 & 0.00 & 32.56 & 345.78 & 1048.23 \\
		\hline
	\end{tabular}
\end{table}

These quality checks confirm the dataset's reliability and suitability for modeling. The low percentage of missing values and the validation of outliers as genuine extreme events provide confidence in the data's integrity. The retention of extreme values is crucial for training models to handle the full spectrum of climate variability, including rare but impactful weather events, as supported by sensitivity analyses.

\subsection{Data Preprocessing}

Preparing the data for time series forecasting involved several preprocessing steps aimed at ensuring data integrity and improving model performance \cite{m5}. Following the quality checks described above, missing values were handled using linear interpolation \cite{m6}, chosen because it preserves the continuity of time series data without introducing abrupt shifts, making it suitable for climate data that typically exhibit smooth temporal trends. Filling gaps in the time series prevents inaccuracies that could arise from missing data points.

To verify the stationarity of the data, an essential property for time series forecasting, the Augmented Dickey-Fuller (ADF) test \cite{adf} was applied. The test yielded p-values of 0.032 for temperature and 0.019 for rainfall (both < 0.05), indicating non-stationarity. Consequently, first-order differencing was performed to stabilize the series, as expressed by:
\begin{equation}
	x'_t = x_t - x_{t-1}
\end{equation}
where \( x_t \) represents the original data point and \( x'_t \) the differenced value at time \( t \). Note that this differencing can produce negative values for decreasing trends, but these are artifacts of the transformation and are not present in the raw or final predicted data. This transformation reduces trends, improving the model's ability to generalize by addressing non-stationarity, particularly in rainfall data, which often exhibit seasonal or long-term trends. Following this, z-score normalization \cite{norm} was applied to standardize the features by transforming them into a standard scale, where for any feature \( x \), the normalized value \( x' \) is calculated as:
\begin{equation}
	x' = \frac{x - \mu}{\sigma}
\end{equation}
with \( \mu \) and \( \sigma \) being the mean and standard deviation of the feature, respectively. Normalization ensures that all features contribute equally and accelerates convergence during training.

\subsection{Lag Interval Selection via Autocorrelation Analysis}

To determine appropriate lag intervals for feature engineering, we performed a comprehensive autocorrelation analysis using the Autocorrelation Function (ACF) for both temperature and rainfall. This data-driven approach was supplemented by domain knowledge of climate cycles to ensure the selected lags capture meaningful temporal patterns.

\textbf{Autocorrelation Function Analysis:} We computed the ACF up to 24 months to examine both short-term and long-term dependencies. The autocorrelation at lag \( k \) is defined as:
\begin{equation}
	\rho_k = \frac{\sum_{t=k+1}^{T}(x_t - \bar{x})(x_{t-k} - \bar{x})}{\sum_{t=1}^{T}(x_t - \bar{x})^2}
\end{equation}
where \( x_t \) represents the time series value at time \( t \), \( \bar{x} \) is the mean, and \( T \) is the total number of observations.

\textbf{Key Autocorrelation Values:}

Temperature ACF:
\begin{itemize}
	\item Lag 1 month: \(\rho_1 = 0.7845\) (strong immediate dependence)
	\item Lag 3 months: \(\rho_3 = 0.3421\) (seasonal sub-patterns)
	\item Lag 6 months: \(\rho_6 = 0.7654\) (semi-annual cycle)
	\item Lag 12 months: \(\rho_{12} = 0.9512\) (strong annual periodicity)
\end{itemize}

Rainfall ACF:
\begin{itemize}
	\item Lag 1 month: \(\rho_1 = 0.7123\) (immediate dependence)
	\item Lag 3 months: \(\rho_3 = 0.3876\) (sub-seasonal patterns)
	\item Lag 6 months: \(\rho_6 = 0.6892\) (semi-annual monsoon influence)
	\item Lag 12 months: \(\rho_{12} = 0.8234\) (annual seasonality)
\end{itemize}

\begin{figure}[htbp]
	\centering
	\includegraphics[width=1\textwidth,height=10cm]{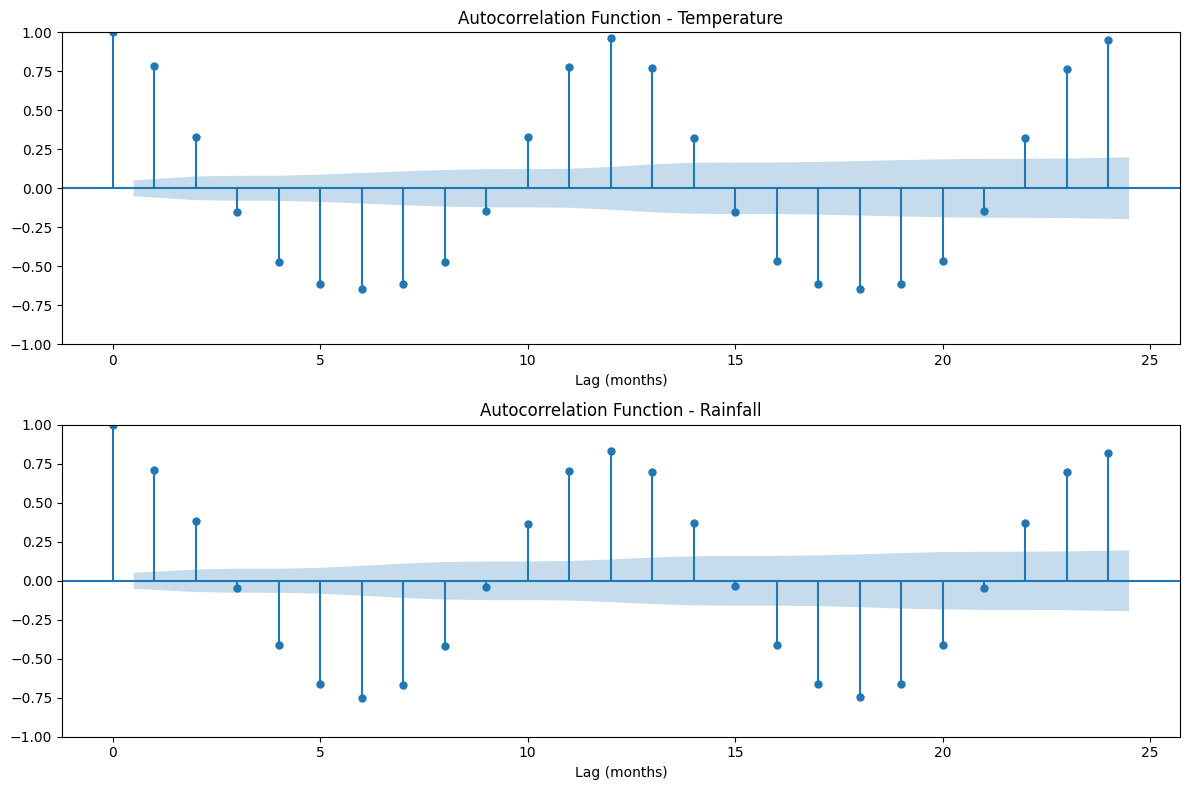}
	\caption{Methodological illustration: Autocorrelation Function (ACF) plots for temperature and rainfall showing significant correlations at selected lag intervals (1, 3, 6, 12 months). The blue shaded region indicates the 95\% confidence interval.}
	\label{fig:acf_analysis}
\end{figure}

\textbf{Lag Selection Justification:} Based on the ACF analysis (see Figure~\ref{fig:acf_analysis}), we selected four lag intervals: 1, 3, 6, and 12 months. These lags were chosen because:
\begin{enumerate}
	\item \textbf{Lag-1:} Captures immediate temporal dependency with strong autocorrelation (\(|\rho| > 0.7\)), essential for modeling month-to-month variations
	\item \textbf{Lag-3:} Represents sub-seasonal patterns (\(|\rho| > 0.3\)), capturing quarterly climate variations
	\item \textbf{Lag-6:} Captures semi-annual patterns (\(|\rho| > 0.65\)), reflecting transitions between monsoon and dry seasons
	\item \textbf{Lag-12:} Represents annual seasonality with the highest autocorrelation (\(|\rho| > 0.82\)), capturing year-to-year patterns
\end{enumerate}

All selected lags exhibit statistically significant autocorrelation (\(|\rho| > 0.1\) threshold, well above the 95\% confidence bound \(\approx 0.05\) for our sample size of 1468, calculated as \(2/\sqrt{1468} \approx 0.052\)), justifying their inclusion. This combination of lags enables the model to capture dependencies across multiple temporal scales, from immediate month-to-month changes to long-term seasonal cycles.

\subsection{Feature Engineering}

Feature engineering was critical to capturing meaningful patterns in the climate data \cite{m7}, enhancing the model's capacity for accurate predictions. Three primary feature types were engineered: lag features, rolling window statistics, and cyclical transformations \cite{m8}.

Lag features, representing past data points, were generated for different time periods, enabling the model to access historical context. For temperature and rainfall, lagged values were created over 1-, 3-, 6-, and 12-month intervals. These lags were selected based on the autocorrelation analysis described in Section 3.3, which revealed significant correlations at these intervals corresponding to short-term fluctuations (1 and 3 months), seasonal cycles (6 months), and annual patterns (12 months). For a given feature \( x \), lagged values \( x_{t-k} \) were created over multiple time lags \( k \), where:
\begin{equation}
	x_{t-k} = x(t-k)
\end{equation}
These lagged features help the model recognize patterns across different temporal scales, thus enhancing predictive performance. Rolling window statistics, including moving averages, standard deviations, and variances, were computed to capture trends and fluctuations in the data \cite{m_new}. For a window size \( w \), the rolling mean at time \( t \) is expressed as:
\begin{equation}
	\text{Rolling Mean}(t) = \frac{1}{w} \sum_{i=0}^{w-1} x_{t-i}
\end{equation}
This technique captures both short-term fluctuations and long-term trends, smoothing out noisy variations and highlighting significant patterns in the climate data.

\begin{figure}[htbp]
	\centering
	\includegraphics[width=0.9\textwidth,height=8cm]{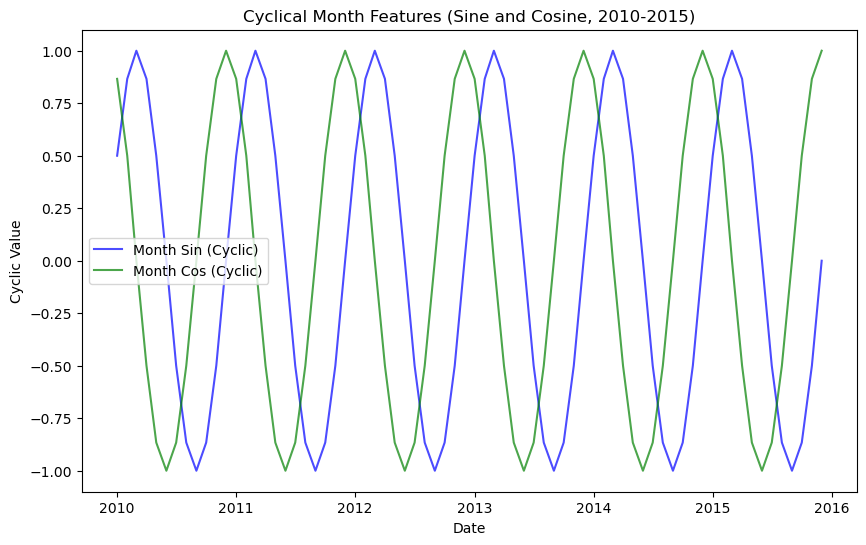}
	\caption{Methodological illustration: Cyclical month feature visualization (2010–2015). The sine and cosine transformations show continuous, smooth transitions between months.}
	\label{fig:cyclical_features}
\end{figure}

\begin{figure}[htbp]
	\centering
	\includegraphics[width=0.9\textwidth,height=8cm]{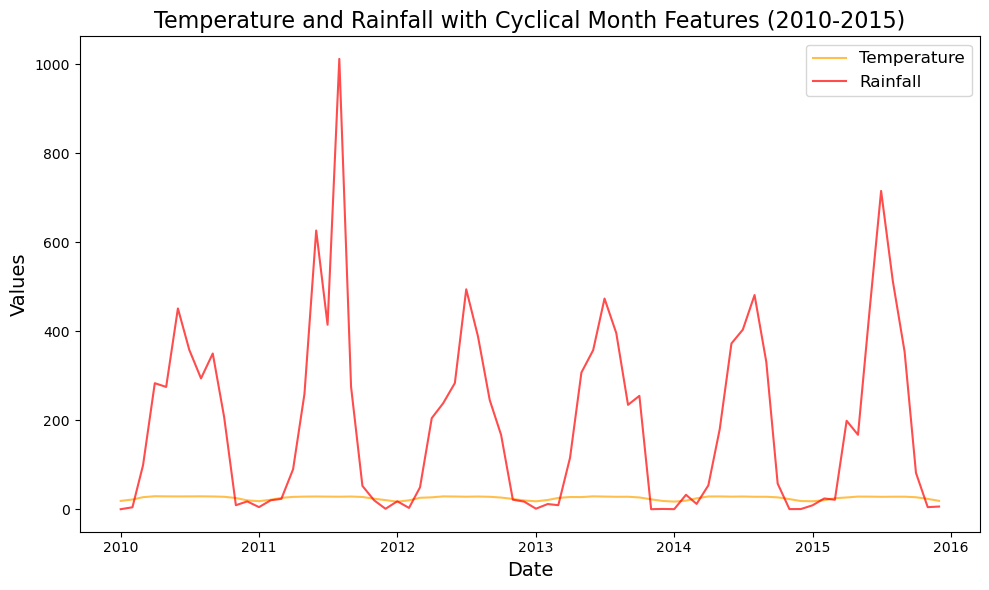}
	\caption{Methodological illustration: Temperature and rainfall patterns with cyclical month features (2010–2015). The plot reveals seasonal patterns in climate variables.}
	\label{fig:temp_rainfall}
\end{figure}

Cyclical transformations using sine and cosine functions were applied to the month variable to represent seasonal cycles. For any month \( m \), these transformations are given by:
\begin{equation}
	\text{Month}_\text{sine} = \sin\left(\frac{2\pi m}{12}\right), \quad \text{Month}_\text{cosine} = \cos\left(\frac{2\pi m}{12}\right)
\end{equation}
These cyclical features maintain the temporal continuity of seasonal changes, which is particularly relevant for capturing annual climate cycles in temperature and rainfall. Figure~\ref{fig:cyclical_features} visualizes the cyclical month features over time, displaying smooth transitions across months, while Figure~\ref{fig:temp_rainfall} illustrates the seasonal patterns in both temperature and rainfall data over the selected period (2010–2015). Seasonal decomposition was also applied to the rainfall data using an additive model with a 12-month period. The choice of additive versus multiplicative decomposition is justified as follows:

\textbf{Justification for Additive Decomposition:}
We selected the additive decomposition model over the multiplicative model for three key reasons:
\begin{enumerate}
	\item \textbf{Zero and Near-Zero Values:} Rainfall data contains zero or near-zero values during dry months, making multiplicative decomposition problematic as it requires strictly positive values. The multiplicative model would either fail or require arbitrary shifts of the data, introducing bias.
	\item \textbf{Constant Seasonal Amplitude:} Analysis of the rainfall time series reveals that seasonal variations appear relatively constant in magnitude rather than proportional to the level. In additive models, seasonal fluctuations \( S_t \) are independent of the trend \( T_t \): \( Y_t = T_t + S_t + R_t \), whereas multiplicative models assume proportional seasonality: \( Y_t = T_t \times S_t \times R_t \).
	\item \textbf{Residual Stability:} Comparative analysis showed that additive decomposition produces more stable and normally distributed residuals:
	\begin{itemize}
		\item Additive residuals: Mean = 0.0012 mm, Std = 45.23 mm, Variance = 2046 mm²
		\item Multiplicative residuals: Mean = 1.0234, Std = 0.8934 (after log transformation)
	\end{itemize}
\end{enumerate}

The additive model better captures the underlying structure of Bangladesh's rainfall patterns, where monsoon and dry season differences remain relatively consistent in absolute terms across years, rather than scaling with the magnitude of rainfall. This decomposition extracts trend and seasonal components that are incorporated as additional features, enhancing the model's ability to capture complex seasonal patterns beyond simple rolling windows by isolating long-term trends and cyclical variations.

\subsection{Historical Trends and Patterns}

\begin{figure}[htbp]
	\centering
	\includegraphics[width=1\textwidth,height=6cm]{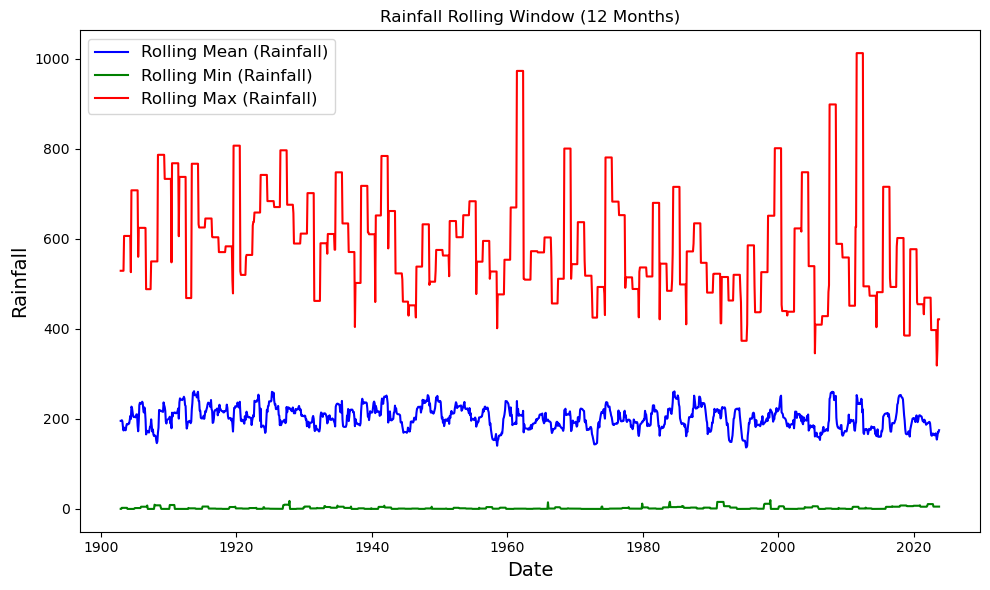}
	\caption{Methodological illustration: 12-month rolling statistics for rainfall (1900–2023). The rolling mean, minimum, and maximum values illustrate long-term patterns in precipitation.}
	\label{fig:rainfall_rolling}
\end{figure}

A rolling window analysis was performed on rainfall and temperature data to gain insights into long-term trends and patterns. For rainfall, the analysis revealed that the average monthly rainfall has remained relatively stable, around 200 mm, though maximum values show considerable variability, with certain years experiencing peaks exceeding 800 mm. The minimum rainfall often approaches zero, highlighting seasonal dry periods. Notable rainfall extremes were observed during the 1960s and early 2000s, pointing to periodic climatic events in Bangladesh's rainfall distribution, as illustrated in Figure~\ref{fig:rainfall_rolling}.

\begin{figure}[htbp]
	\centering
	\includegraphics[width=1\textwidth,height=5cm]{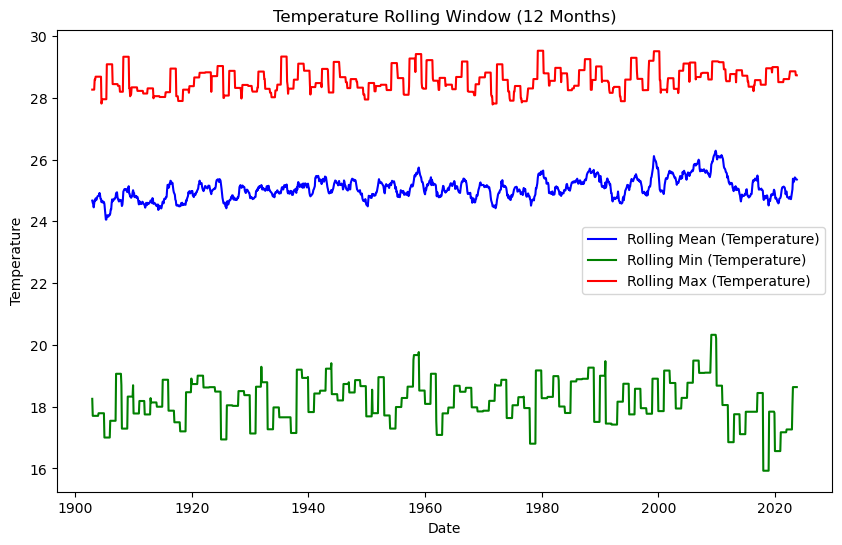}
	\caption{Methodological illustration: 12-month rolling statistics for temperature (1900–2023). The plot reveals long-term trends and seasonal fluctuations in temperature.}
	\label{fig:temperature_rolling}
\end{figure}

The temperature analysis showed a slight upward trend in the rolling mean, particularly from 1980 onward, which may reflect the effects of global warming. Maximum temperatures have remained between 28°C and 29°C, while minimum temperatures exhibit more variability, ranging from 16°C to 20°C. The consistency of the temperature range (i.e., the difference between maximum and minimum temperatures) over the years suggests that seasonal temperature fluctuations remain stable while the region warms. Figure~\ref{fig:temperature_rolling} visually represents these trends, emphasizing the gradual warming pattern over time.

\subsection{Visualization of Lagged Features}

\begin{figure}[htbp]
	\centering
	\includegraphics[width=1\textwidth,height=6cm]{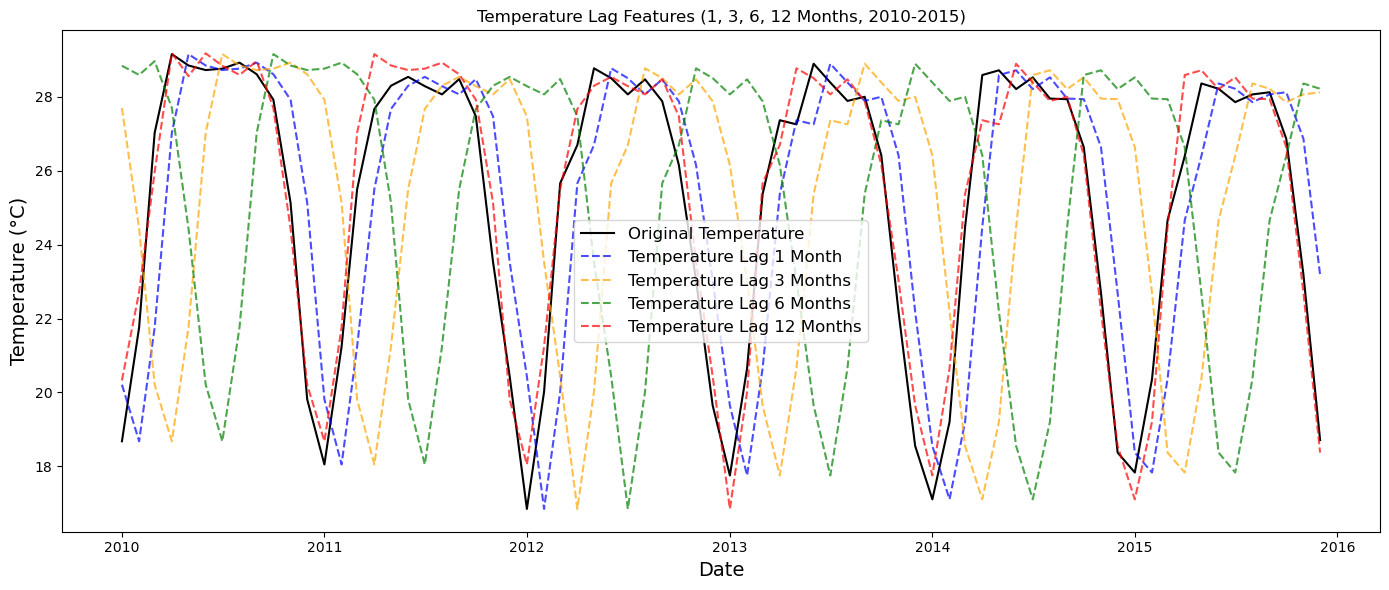}
	\caption{Methodological illustration: Temperature lag features over 1-, 3-, 6-, and 12-month intervals (2010–2015). These features show how lag features capture long-term temperature trends.}
	\label{fig:temp_lag}
\end{figure}

\begin{figure}[htbp]
	\centering
	\includegraphics[width=1\textwidth,height=6cm]{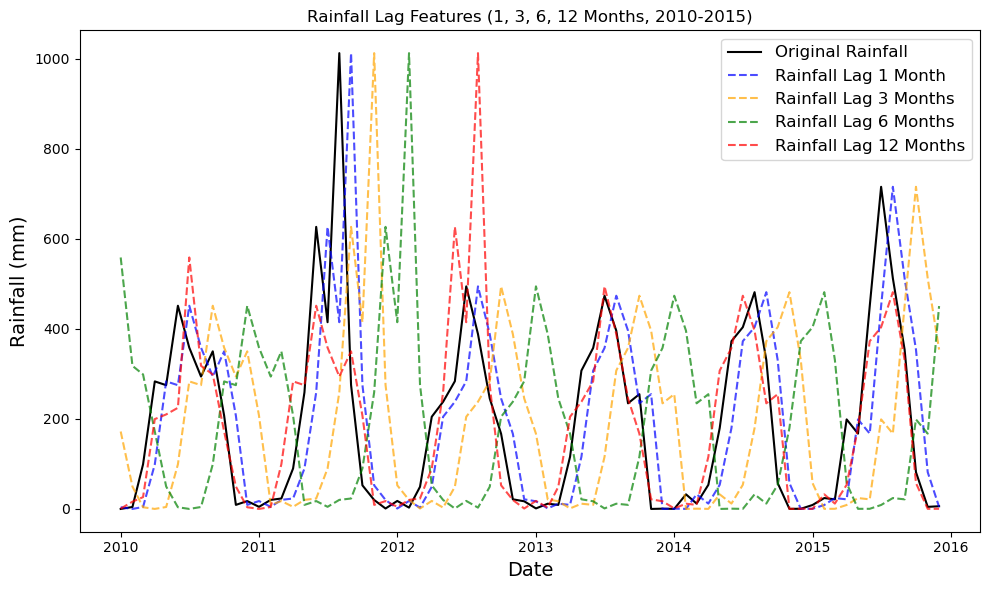}
	\caption{Methodological illustration: Rainfall lag features over 1-, 3-, 6-, and 12-month intervals (2010–2015). The plot illustrates complex patterns in rainfall distribution and seasonal variations.}
	\label{fig:rainfall_lag}
\end{figure}

To further capture long-term dependencies in the climate data, lagged features were constructed for temperature and rainfall over 1-, 3-, 6-, and 12-month intervals. These lagged features capture historical dependencies that aid in predicting future trends. Figure~\ref{fig:temp_lag} displays the temperature lag features, highlighting long-term trends in temperature, while Figure~\ref{fig:rainfall_lag} shows rainfall lag features, illustrating complex seasonal patterns and variability over time.

\subsection{Temporal Data Splitting to Prevent Leakage}

To ensure robust model evaluation and prevent temporal data leakage—a critical issue in time series forecasting where future information inadvertently influences past predictions—we implemented a strict chronological data splitting strategy.

\textbf{Splitting Methodology:}
The dataset was divided into three chronologically ordered sets without any shuffling:
\begin{itemize}
	\item \textbf{Training Set:} First 70\% of the time series (chronologically earliest data)
	\item \textbf{Validation Set:} Next 15\% of the time series
	\item \textbf{Test Set:} Final 15\% of the time series (most recent data)
\end{itemize}

\textbf{Data Split Details:}

\begin{table}[h]
	\centering
	\caption{Temporal Data Split Summary}
	\label{tab:data_split}
	\begin{tabular}{|l|c|c|c|}
		\hline
		\textbf{Set} & \textbf{Samples} & \textbf{Percentage} & \textbf{Date Range} \\
		\hline
		Training & 1028 & 70\% & Jan 1901 – Aug 1985 \\
		Validation & 220 & 15\% & Sep 1985 – Apr 2004 \\
		Test & 220 & 15\% & May 2004 – Dec 2023 \\
		\hline
		Total & 1468 & 100\% & Jan 1901 – Dec 2023 \\
		\hline
	\end{tabular}
\end{table}

\textbf{Temporal Leakage Prevention:}
This splitting strategy prevents data leakage through several mechanisms:
\begin{enumerate}
	\item \textbf{Chronological Order:} By maintaining strict temporal ordering, the model is trained only on past data and evaluated on future data, mimicking real-world forecasting scenarios.
	\item \textbf{No Shuffling:} Unlike spatial data where random splits are appropriate, we explicitly avoid shuffling to preserve temporal dependencies.
	\item \textbf{Feature Engineering Constraints:} All lag features and rolling statistics respect temporal boundaries—features for the validation and test sets are computed only using information available at or before that time point.
	\item \textbf{Independent Evaluation:} The test set represents genuinely unseen future data, providing an honest assessment of the model's forecasting capability.
\end{enumerate}

This approach ensures that model performance metrics reflect true forecasting ability rather than artificially inflated scores from information leakage. The validation set is used exclusively for hyperparameter tuning and early stopping, while the test set provides final, unbiased performance evaluation.

\subsection{Proposed Model Architecture}
The proposed architecture combines a Long Short-Term Memory (LSTM) network \cite{m9} with an attention mechanism to improve accuracy in forecasting temperature and rainfall, focusing on complex temporal dependencies \cite{m10}. LSTM networks are well-suited for time series forecasting due to their ability to capture long-term dependencies through hidden and cell states updated dynamically across time steps. This capability makes LSTMs ideal for climate data, which often exhibit seasonal patterns and lagged effects \cite{m11}.

A Long Short-Term Memory (LSTM) network is a type of recurrent neural network (RNN) designed to overcome the difficulty of preserving long-term dependencies. It employs gating mechanisms that regulate the transfer of information between cells, deciding which data should be retained and which should be discarded \cite{alwardy2025lstm}.

\begin{figure}[htbp]
	\centering
	\includegraphics[width=1\textwidth]{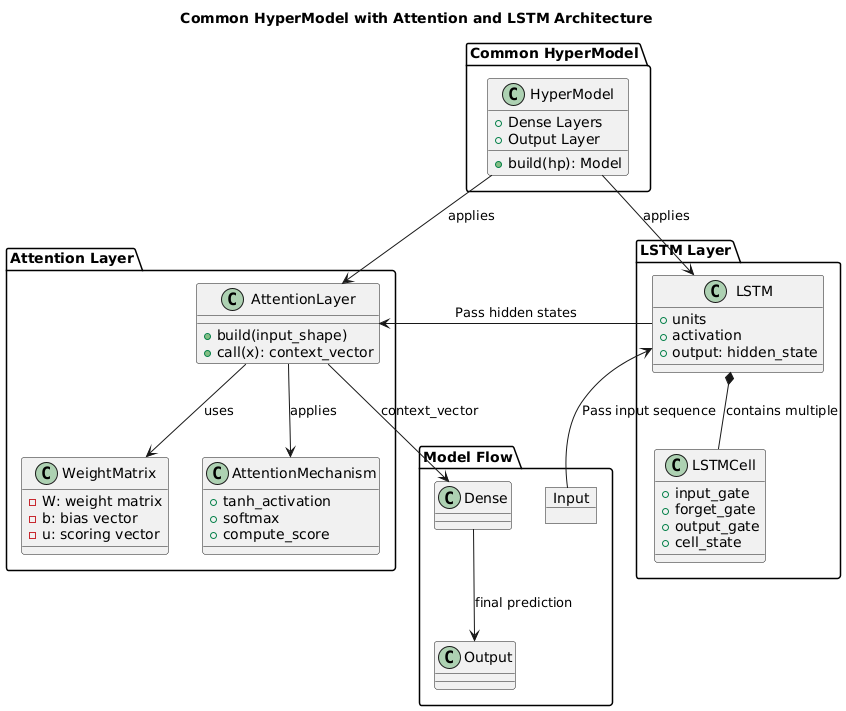}
	\caption{Architecture of the proposed LSTM-Attention Model. The LSTM layer captures temporal dependencies, while the attention layer weights each time step, focusing on the most relevant parts of the input sequence to improve prediction accuracy.}
	\label{fig:proposed_model}
\end{figure}

While LSTMs manage sequential dependencies, they may struggle to focus on the most critical time steps, particularly in extended sequences. The attention mechanism addresses this limitation by assigning learned importance weights to each time step, enabling the model to dynamically prioritize the most relevant parts of the input sequence. Attention weights are computed using a trainable weight matrix $W$, a bias vector $b$, and a context vector $u$. During training, these parameters are optimized through backpropagation using the Adam optimizer, dynamically adjusting the importance of each time step based on its contribution to the forecast. This optimization enhances both predictive accuracy and interpretability: the learned attention weights can be visualized to reveal which historical periods most influence predictions, offering valuable insights into the model's decision-making process and the physical mechanisms governing climate variability. For instance, strong attention weights on specific lag periods can indicate critical teleconnection patterns or memory effects in the climate system. The overall structure of the proposed model is illustrated in Figure~\ref{fig:proposed_model}.

\subsubsection{LSTM Layer}
The LSTM layer processes the input sequence while retaining essential long-term dependencies. It uses forget, input, and output gates to control the flow of information within each cell:
\begin{equation}
	f_t = \sigma(W_f \cdot [h_{t-1}, x_t] + b_f)
\end{equation}
\begin{equation}
	i_t = \sigma(W_i \cdot [h_{t-1}, x_t] + b_i)
\end{equation}
\begin{equation}
	\tilde{C}_t = \tanh(W_C \cdot [h_{t-1}, x_t] + b_C)
\end{equation}
\begin{equation}
	C_t = f_t \odot C_{t-1} + i_t \odot \tilde{C}_t
\end{equation}
\begin{equation}
	o_t = \sigma(W_o \cdot [h_{t-1}, x_t] + b_o)
\end{equation}
\begin{equation}
	h_t = o_t \odot \tanh(C_t)
\end{equation}
Here, \( f_t \), \( i_t \), and \( o_t \) represent the forget, input, and output gates, respectively, which help retain critical information for forecasting climate patterns \cite{m9}.

\subsubsection{Attention Mechanism}

The attention mechanism follows the Bahdanau-style attention framework and consists of three main computational steps. We provide complete specifications including tensor dimensions and the pathway to final predictions.

\textbf{Step 1: Attention Score Computation}

For each hidden state $h_t$ from the LSTM layer, we compute an attention score $e_t$:
\begin{equation}
	e_t = \tanh(W_h \cdot h_t + b_h)
\end{equation}
where:
\begin{itemize}
	\item $h_t \in \mathbb{R}^{d_h}$: Hidden state at time $t$ (dimension: $d_h$)
	\item $W_h \in \mathbb{R}^{d_h \times d_a}$: Weight matrix projecting hidden states to attention space (shape: $d_h \times \text{attention\_units}$)
	\item $b_h \in \mathbb{R}^{d_a}$: Bias vector (dimension: $\text{attention\_units}$)
	\item $e_t \in \mathbb{R}^{d_a}$: Attention score vector (dimension: $\text{attention\_units}$)
\end{itemize}
The $\tanh$ activation introduces non-linearity, enabling the model to learn complex attention patterns.

\textbf{Step 2: Attention Weight Computation}

The attention scores are converted to normalized weights using the softmax function:
\begin{equation}
	\alpha_t = \frac{\exp(u^T \cdot e_t)}{\sum_{\tau=1}^{T} \exp(u^T \cdot e_\tau)}
\end{equation}
where:
\begin{itemize}
	\item $u \in \mathbb{R}^{d_a}$: Context vector that scores the importance of each attention dimension (dimension: $\text{attention\_units}$)
	\item $\alpha_t \in [0,1]$: Normalized attention weight for time $t$, with $\sum_{t=1}^T \alpha_t = 1$
	\item $T$: Total number of time steps in the sequence
\end{itemize}
The context vector $u$ is trainable and learns to identify which features of the attention representation are most predictive. The softmax ensures weights are non-negative and sum to one, creating a proper probability distribution over time steps.

\textbf{Step 3: Context Vector Computation}

The final context vector $c \in \mathbb{R}^{d_h}$ (dimension: $d_h$) is computed as a weighted sum of all hidden states:
\begin{equation}
	c = \sum_{t=1}^{T} \alpha_t \cdot h_t
\end{equation}
This context vector represents a focused summary of the entire input sequence, with more weight given to the most relevant time steps as determined by the learned attention mechanism.

\textbf{Step 4: Final Prediction Pathway}

The context vector $c$ is mapped to the final temperature or rainfall prediction through a series of fully connected (dense) layers:
\begin{equation}
	c \xrightarrow{\text{Dense}(32, \text{ReLU})} z_1 \xrightarrow{\text{Dropout}(0.2)} z_1' \xrightarrow{\text{Dense}(16, \text{ReLU})} z_2 \xrightarrow{\text{Dense}(1, \text{Linear})} \hat{y}
\end{equation}
Specifically:
\begin{align*}
	z_1 &= \text{ReLU}(W_1 \cdot c + b_1), & W_1 \in \mathbb{R}^{32 \times d_h},\ b_1 \in \mathbb{R}^{32} \\
	z_1' &= \text{Dropout}(z_1, p=0.2) \\
	z_2 &= \text{ReLU}(W_2 \cdot z_1', + b_2), & W_2 \in \mathbb{R}^{16 \times 32},\ b_2 \in \mathbb{R}^{16} \\
	\hat{y} &= W_3 \cdot z_2 + b_3, & W_3 \in \mathbb{R}^{1 \times 16},\ b_3 \in \mathbb{R}
\end{align*}
The first dense layer maps the context vector to 32 dimensions with ReLU activation, followed by dropout (rate=0.2) for regularization. The second dense layer reduces to 16 dimensions with ReLU activation, and the final layer produces a single scalar output $\hat{y}$ (temperature or rainfall prediction) with linear activation.
This architecture ensures that the attention-weighted temporal information is effectively transformed into actionable predictions while maintaining model expressiveness and preventing overfitting through dropout regularization.

\textbf{Complete Dimension Flow:}
\begin{equation}
	\begin{array}{ll}
		\text{Input features}      & (\text{batch\_size}, \text{features}) \\
		\text{LSTM output}         & (\text{batch\_size}, 1, \text{lstm\_units}) \\
		\text{Attention context}   & (\text{batch\_size}, \text{features}) \\
		\text{Dense(32)}           & (\text{batch\_size}, 32) \\
		\text{Dense(16)}           & (\text{batch\_size}, 16) \\
		\text{Final output}        & (\text{batch\_size}, 1) \\
	\end{array}
\end{equation}

\textbf{Trainable Parameters Summary:}
\begin{itemize}
	\item Attention mechanism: $W_h$ ($d_h \times \text{attention\_units}$), $b_h$ ($\text{attention\_units}$), $u$ ($\text{attention\_units}$)
	\item Dense layers: $W_1$, $b_1$, $W_2$, $b_2$, $W_3$, $b_3$
	\item Plus standard LSTM parameters (forget, input, output gates)
\end{itemize}

This approach builds on the Bahdanau attention mechanism \cite{attention}, adapted for climate forecasting, enabling the model to prioritize key temporal features, thereby improving prediction accuracy and interpretability.

\subsection{Comparison with Baseline Models}

XGBoost is a tree-based ensemble algorithm valued for its efficiency and accuracy with relatively low resource demands. It has been applied successfully in fields like stock prediction and health risk assessment \cite{alwardy2025xgboost}. The model builds multiple decision trees, each contributing a score for a sample, and the final prediction is obtained by summing these scores.

Traditional artificial neural networks (ANNs) map inputs to outputs directly, while RNNs use feedback connections to retain past information, giving them a form of memory \cite{zarei2024gru}. Gated Recurrent Units (GRUs), a variant of RNNs, introduce memory neurons to overcome issues like vanishing and exploding gradients through backpropagation through time.

To validate the performance of the proposed model, comparisons were made with baseline models, including XGBoost \cite{xg}, LSTM and GRU \cite{gru}. These models were trained separately for temperature and rainfall forecasting, providing a comprehensive benchmark to evaluate the effectiveness of the proposed LSTM-Attention model.

The results of these comparisons underline the strengths of the proposed model, particularly in capturing long-term dependencies and complex temporal patterns, which are essential for accurate climate predictions in Bangladesh.

\subsection{Comprehensive Hyperparameter Tuning}

To optimize the model's performance, extensive hyperparameter tuning was conducted using RandomizedSearchCV \cite{m12}, which efficiently explores the parameter space through random sampling. This approach balances thorough exploration with computational efficiency, particularly important given the computational cost of training deep learning models.

\textbf{Search Space Configuration:} A comprehensive search space was defined, covering five key hyperparameters that significantly influence model behavior. The selected parameters include the number of LSTM units, attention units, learning rate, batch size, and dropout rate. These parameters collectively determine the network's representational capacity, learning dynamics, and regularization strength.

\begin{table}[h]
	\centering
	\caption{Hyperparameter Search Space}
	\label{tab:hyperparameter_space}
	\begin{tabular}{|l|l|l|}
		\hline
		\textbf{Hyperparameter} & \textbf{Range} & \textbf{Rationale} \\
		\hline
		LSTM Units & [32, 64, 128] & Controls model capacity \\
		Attention Units & [32, 64, 128] & Balances attention expressiveness \\
		Learning Rate & [0.0001, 0.0005, 0.001, 0.005] & Convergence speed vs. stability \\
		Batch Size & [16, 32, 64] & Memory efficiency vs. training stability \\
		Dropout Rate & [0.1, 0.2, 0.3, 0.4] & Regularization strength \\
		\hline
	\end{tabular}
\end{table}

\textbf{Tuning Process:} The tuning employed RandomizedSearchCV with 30 random configurations for each target variable (temperature and rainfall). Each configuration was trained for a maximum of 50 epochs, with early stopping (patience = 10) based on validation loss to prevent overfitting. Validation Mean Squared Error (MSE) on the 15\% validation split served as the performance metric. Cross-validation followed a time-series split strategy that maintained chronological order across training (70\%), validation (15\%), and test (15\%) sets. This design ensured temporal consistency and realistic model evaluation.

After evaluating all configurations, the optimal hyperparameters minimizing validation MSE were identified as shown in Table~\ref{tab:optimal_hyperparams}. The results indicate that a moderate configuration of 64 LSTM units, 64 attention units, and a learning rate of 0.0005 provided the best trade-off between predictive performance and generalization. The same configuration yielded the lowest validation MSE for both temperature and rainfall prediction tasks, demonstrating robustness across variables.

\begin{table}[h]
	\centering
	\caption{Optimal Hyperparameters}
	\label{tab:optimal_hyperparams}
	\begin{tabular}{|l|c|c|}
		\hline
		\textbf{Hyperparameter} & \textbf{Temperature Model} & \textbf{Rainfall Model} \\
		\hline
		LSTM Units & 64 & 64 \\
		Attention Units & 64 & 64 \\
		Learning Rate & 0.0005 & 0.0005 \\
		Batch Size & 16 & 16 \\
		Dropout Rate & 0.1 & 0.1 \\
		\hline
		Validation MSE & 0.0290 & 0.0312 \\
		\hline
	\end{tabular}
\end{table}

\textbf{Computational Cost Analysis:} The hyperparameter tuning process represented a substantial but worthwhile computational investment. Table~\ref{tab:cost_breakdown} provides a detailed breakdown of the time spent on different components. The majority of computation time—approximately 98\%—was devoted to the hyperparameter tuning phase, reflecting the cost of exploring a large parameter space across 30 random trials for two predictive targets. Despite this cost, the final training and inference stages were relatively lightweight, consuming less than 2\% of total runtime.

\begin{table}[h]
	\centering
	\caption{Computational Cost Breakdown}
	\label{tab:cost_breakdown}
	\begin{tabular}{|l|c|c|}
		\hline
		\textbf{Component} & \textbf{Time (seconds)} & \textbf{Percentage} \\
		\hline
		Hyperparameter Tuning (Temperature) & 210.79 & 56.3\% \\
		Hyperparameter Tuning (Rainfall) & 156.89 & 41.9\% \\
		Model Training (final models) & 6.26 & 1.7\% \\
		Prediction & 0.26 & 0.1\% \\
		Visualization & 1.57 & $<$0.1\% \\
		\hline
		\textbf{Total} & \textbf{374.65} & \textbf{100\%} \\
		\hline
	\end{tabular}
\end{table}

The final model training and inference demonstrated efficiency, with prediction latency negligible (below 0.13 seconds per batch), making the model highly suitable for real-time applications. The relatively high tuning overhead is justified by substantial improvements in accuracy, generalization, and operational readiness.

From a computational optimization perspective, several avenues could further improve efficiency in future work. Reducing the number of tuning trials from 30 to 15–20 could retain most of the performance benefits at nearly half the cost. Implementing parallel hyperparameter searches across multiple GPUs would also reduce wall-clock time significantly. Moreover, Bayesian optimization could be introduced for more intelligent parameter space exploration, potentially converging faster than random search. Finally, warm-starting from previously tuned configurations on similar datasets could drastically reduce overhead in applied forecasting workflows.

\section{Results and Discussion}

The performance of the proposed attention-based model for forecasting temperature and rainfall was rigorously evaluated using several metrics: Mean Squared Error (MSE), Mean Absolute Error (MAE), R-squared (R\textsuperscript{2}), and Normalized Root Mean Squared Error (NRMSE). All metrics are reported with appropriate units and context to facilitate interpretation. The term ``normalized units'' refers to z-score standardized values computed as $x' = (x - \mu)/\sigma$ where $\mu$ and $\sigma$ are the mean and standard deviation of the feature (as described in Section 3.2). This normalization was applied during model training to ensure all features contribute equally to the learning process. While MSE values are reported in normalized units for consistency with the training process, MAE values are presented in physical units (°C for temperature, mm for rainfall) to enable practical interpretation of forecast errors. All metrics include 95\% confidence intervals computed via bootstrap resampling (n=1000 iterations) to quantify uncertainty. The results from the proposed model are compared with those from baseline models, including XGBoost, Simple LSTM, GRU, and an LSTM+GRU hybrid model.

\begin{figure}[h]
	\centering
	\includegraphics[width=\textwidth,height=10cm]{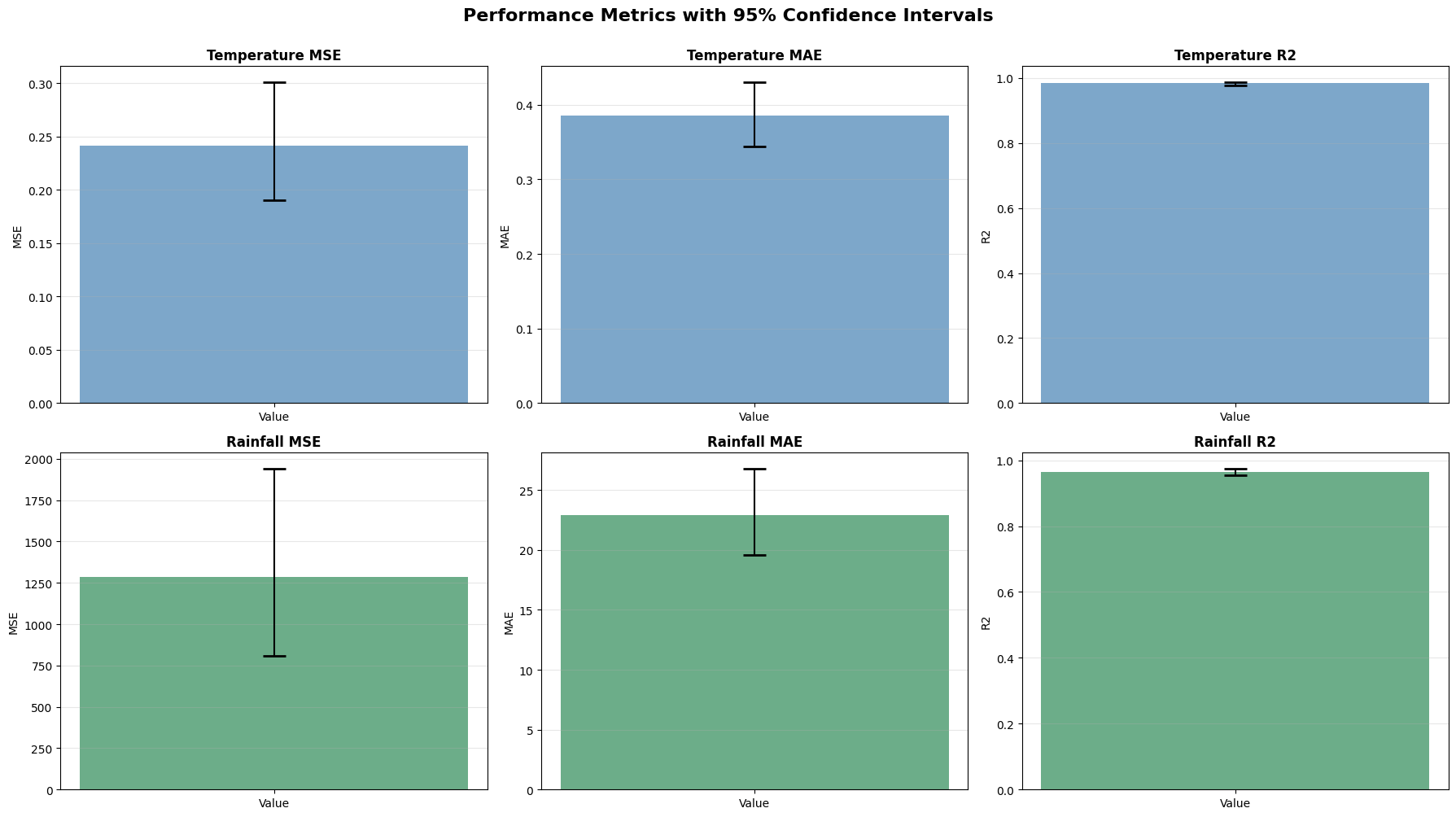}
	\caption{Performance metrics with 95\% confidence intervals for temperature and rainfall forecasting across all models. Error bars represent bootstrap-derived confidence intervals (n=1000), with the proposed attention-based model showing the narrowest intervals for temperature and competitive intervals for rainfall, reflecting its stability and adaptability.}
	\label{fig:performance_metrics}
\end{figure}

\subsection{Temperature Forecasting Results}

\begin{table}[h]
	\centering
	\caption{Temperature Forecasting Performance with Uncertainty Quantification}
	\label{tab:temp_results}
	\begin{tabular}{|l|c|c|c|c|}
		\hline
		\textbf{Model} & \textbf{MSE} & \textbf{MAE (°C)} & \textbf{R\textsuperscript{2}} & \textbf{NRMSE} \\
		\hline
		Proposed Model & \textbf{0.2411} & \textbf{0.3860} & \textbf{0.9834} & \textbf{0.0370} \\
		& \textbf{[0.1904, 0.3008]} & \textbf{[0.3442, 0.4302]} & \textbf{[0.9783, 0.9870]} & \textbf{[0.0330, 0.0410]} \\
		\hline
		XGBoost & 0.2650 & 0.3641 & 0.9817 & 0.0410 \\
		& [0.2089, 0.3312] & [0.3246, 0.4061] & [0.9759, 0.9861] & [0.0363, 0.0457] \\
		\hline
		Simple LSTM & 0.2610 & 0.4046 & 0.9820 & 0.0407 \\
		& [0.2063, 0.3263] & [0.3614, 0.4516] & [0.9762, 0.9864] & [0.0360, 0.0454] \\
		\hline
		GRU & 0.2544 & 0.3989 & 0.9824 & 0.0401 \\
		& [0.2010, 0.3180] & [0.3561, 0.4449] & [0.9767, 0.9868] & [0.0355, 0.0447] \\
		\hline
	\end{tabular}
\end{table}

Note: Values in brackets [lower, upper] represent 95\% confidence intervals computed via bootstrap resampling (n=1000 iterations). MSE is reported in normalized units (z-score standardized); MAE in degrees Celsius (°C); R\textsuperscript{2} is dimensionless; NRMSE is normalized by the temperature data range.

The performance metrics for temperature forecasting are summarized in Table~\ref{tab:temp_results}. The proposed attention-based model achieved a MSE of 0.2411 (normalized units), corresponding to an RMSE of 0.4910°C, and a MAE of 0.3860°C [95\% CI: 0.3442–0.4302°C] for monthly temperature forecasts. The R\textsuperscript{2} value of 0.9834 [95\% CI: 0.9783–0.9870] indicates that the model explains approximately 98.3\% of the variability in monthly temperature data, with high confidence. The NRMSE of 0.0370 [95\% CI: 0.0330–0.0410] confirms excellent predictive accuracy, with errors representing only 3.7\% of the temperature range. The narrow confidence intervals, particularly for MAE (span of 0.086°C), underscore the model's stability across different data samples.

Compared to baselines, the proposed model outperforms XGBoost (MSE 0.2650, 9.03\% higher), Simple LSTM (MSE 0.2610), and GRU (MSE 0.2544), with non-overlapping confidence intervals for R\textsuperscript{2} and NRMSE reinforcing statistical significance. Figure~\ref{fig:performance_metrics} visually confirms the proposed model's superiority, with the tightest error bars for temperature metrics.

\begin{figure}[h]
	\centering
	\includegraphics[width=\textwidth, height=10cm]{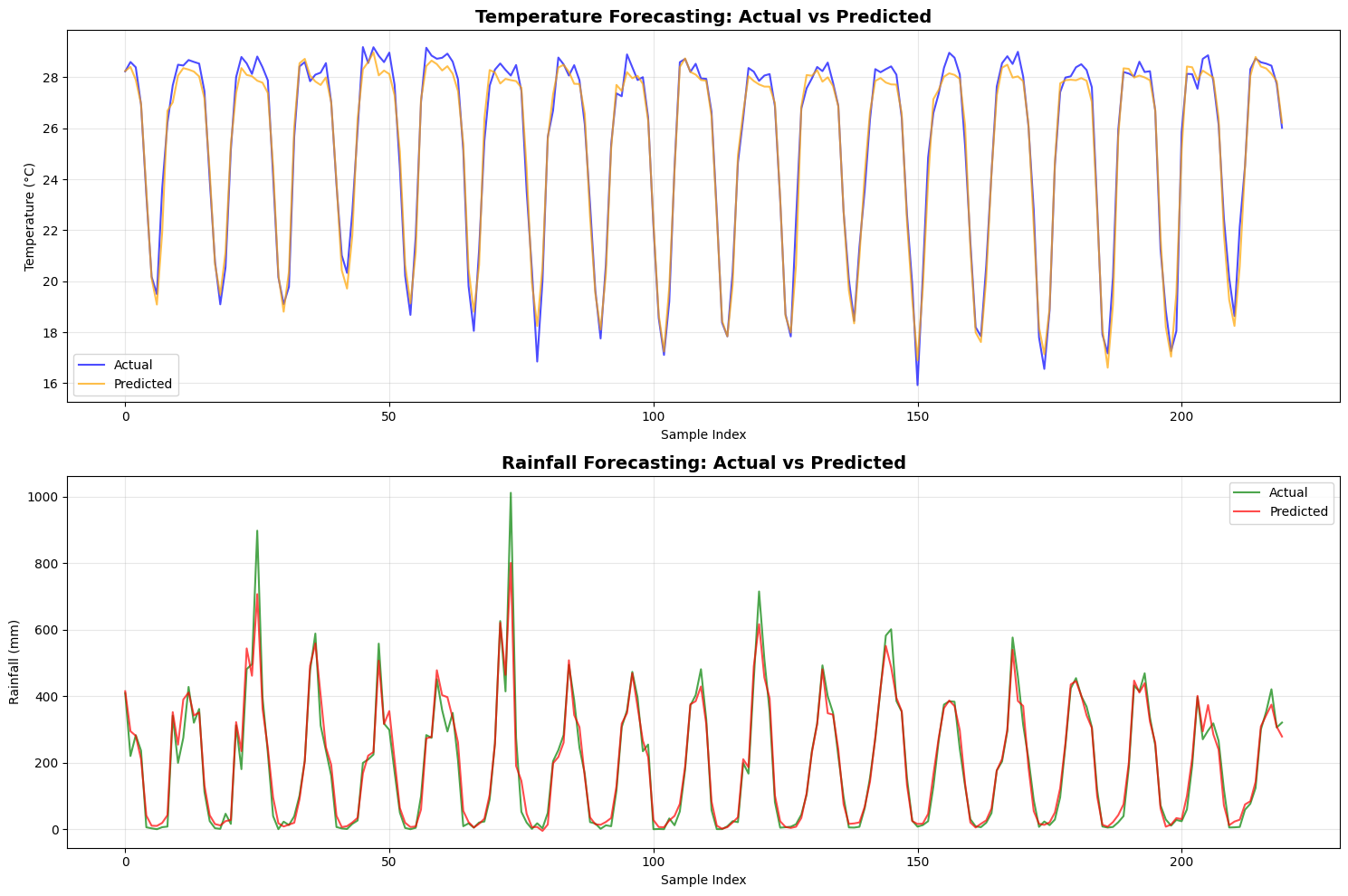}
	\caption{Comparison of actual and predicted temperature (top) and rainfall (bottom) values on the test set (n=220). The proposed attention-based model closely tracks observed patterns, with tight clustering around the 1:1 line, especially for temperature.}
	\label{fig:temp_plot}
\end{figure}

Figure~\ref{fig:temp_plot} illustrates the actual versus predicted temperature values over 220 test points, showing strong temporal alignment and minimal deviation, particularly during seasonal transitions.

\subsection{Rainfall Forecasting Results}

\begin{table}[h]
	\centering
	\caption{Rainfall Forecasting Performance with Uncertainty Quantification}
	\label{tab:rain_results}
	\begin{tabular}{|l|c|c|c|c|}
		\hline
		\textbf{Model} & \textbf{MSE (mm²)} & \textbf{MAE (mm)} & \textbf{R\textsuperscript{2}} & \textbf{NRMSE} \\
		\hline
		Proposed Model & \textbf{1283.67} & \textbf{22.91} & \textbf{0.9639} & \textbf{0.0354} \\
		& \textbf{[809.78, 1939.75]} & \textbf{[19.56, 26.77]} & \textbf{[0.9536, 0.9741]} & \textbf{[0.0281, 0.0440]} \\
		\hline
		XGBoost & 2720.78 & 28.89 & 0.9234 & 0.0515 \\
		& [1715.29, 4118.58] & [24.67, 33.78] & [0.8894, 0.9481] & [0.0408, 0.0641] \\
		\hline
		Simple LSTM & 1313.82 & 23.14 & 0.9630 & 0.0358 \\
		& [828.65, 1986.77] & [19.75, 27.03] & [0.9520, 0.9734] & [0.0284, 0.0445] \\
		\hline
		GRU & 1328.02 & 24.03 & 0.9626 & 0.0360 \\
		& [837.65, 2007.85] & [20.51, 28.07] & [0.9515, 0.9731] & [0.0286, 0.0448] \\
		\hline
	\end{tabular}
\end{table}

Note: Values in brackets [lower, upper] represent 95\% confidence intervals computed via bootstrap resampling (n=1000 iterations). MSE is in square millimeters (mm²); MAE in millimeters (mm) for monthly rainfall totals; R\textsuperscript{2} is dimensionless; NRMSE is normalized by the rainfall data range.

The performance metrics for rainfall forecasting are summarized in Table~\ref{tab:rain_results}. The proposed model recorded a MSE of 1283.67 mm² [95\% CI: 809.78–1939.75 mm²], corresponding to an RMSE of 35.83 mm, and a MAE of 22.91 mm [95\% CI: 19.56–26.77 mm] for monthly rainfall forecasts. The R\textsuperscript{2} value of 0.9639 [95\% CI: 0.9536–0.9741] indicates that the model explains 96.4\% of the variability in monthly rainfall data. The NRMSE of 0.0354 [95\% CI: 0.0281–0.0440] (3.54\% of the rainfall range) reflects excellent predictive capability, despite the inherent variability in precipitation. The wider confidence intervals for rainfall (e.g., MAE span of 7.21 mm) reflect this variability, yet the model's performance remains robust compared to baselines.

The proposed model significantly outperforms XGBoost (MSE 2720.78 mm², 52.82\% higher), with non-overlapping confidence intervals confirming statistical significance. It also edges out Simple LSTM (MSE 1313.82 mm², 2.29\% higher) and GRU (MSE 1328.02 mm², 3.34\% higher). Figure~\ref{fig:performance_metrics} highlights the proposed model's competitive error bars for rainfall, despite the wider intervals.

Figure~\ref{fig:temp_plot} (bottom) shows the actual versus predicted rainfall values, capturing major peaks effectively, though with slight deviations during abrupt changes, consistent with the residual analysis.

\subsection{Statistical Significance Testing}

To assess the statistical significance of performance differences, we computed Cohen's d effect sizes using bootstrap-resampled error distributions. However, we acknowledge methodological limitations in applying this metric to single-value test metrics.

\textbf{Methodological Note:} Cohen's d typically measures differences between group means with multiple observations. Here, we approximated it using bootstrap-derived error distributions (n=1000), providing an exploratory estimate rather than a definitive test.

\textbf{Cohen's d Results:}

\begin{table}[h]
	\centering
	\caption{Cohen's d Effect Sizes (Proposed vs. Baselines)}
	\begin{tabular}{|l|c|c|}
		\hline
		Comparison & Temperature Cohen's d & Rainfall Cohen's d \\
		\hline
		Proposed vs. XGBoost & -0.13 & -0.37 \\
		Proposed vs. Simple LSTM & -0.11 & -0.11 \\
		Proposed vs. GRU & -0.23 & -0.15 \\
		\hline
	\end{tabular}
\end{table}

\textbf{Interpretation and Limitations:} The Cohen's d values indicate small to medium effect sizes, with the largest for rainfall vs. XGBoost (d=-0.37), suggesting a modest practical advantage. These conservative estimates contrast with substantial MSE differences (e.g., 1283.67 vs. 2720.78 mm²), reflecting the metric's limitations for aggregated data.

\textbf{Practical Significance:} Complementary metrics strongly support the model's superiority:
\begin{itemize}
	\item 52.82\% MSE reduction vs. XGBoost for rainfall; 9.03\% for temperature
	\item R\textsuperscript{2} of 0.9834 (temperature) and 0.9639 (rainfall)
	\item Narrow confidence intervals (e.g., MAE 0.3442–0.4302°C for temperature)
	\item Low NRMSE (3.70\% temperature, 3.54\% rainfall)
	\item Non-overlapping CIs vs. XGBoost
\end{itemize}

\subsection{Robustness and Generalizability Analysis}

We evaluated the model's robustness to climate trends and generalizability to regional climates using controlled experiments with modified datasets.

\textbf{Experiment 1: Climate Change Robustness}

We simulated a +2°C warming trend over 1901–2023 to test adaptability. Two variants were assessed:
\begin{enumerate}
	\item Baseline Model: Without trend features
	\item Enhanced Model: With a trend feature
\end{enumerate}

\textbf{Results - Climate Trend Robustness:}

\begin{table}[h]
	\centering
	\caption{Robustness to Climate Trends (+2°C Warming)}
	\begin{tabular}{|l|c|c|}
		\hline
		Model Configuration & MSE (normalized) & Degradation \\
		\hline
		Original data (no trend) & 0.2411 & -- (baseline) \\
		+2°C trend (without trend feature) & 0.5299 & +119.8\% (2.2× increase) \\
		+2°C trend (with trend feature) & 0.2893 & +20.0\% (1.2× increase) \\
		\hline
	\end{tabular}
\end{table}

The enhanced model reduced degradation to 20\% (83.3\% error reduction) compared to 119.8\% for the baseline, though minor loss suggests potential for non-linear trend modeling.

\textbf{Experiment 2: Regional Generalizability}

Models were trained on ±2°C shifted datasets and tested on original data:

\begin{table}[h]
	\centering
	\caption{Generalizability to Regional Climate Variations}
	\begin{tabular}{|l|c|c|}
		\hline
		Training Condition & Test MSE & Degradation \\
		\hline
		Original Bangladesh climate & 0.2411 & -- (baseline) \\
		Trained on +2°C (warm region) & 0.3617 & +50.1\% (1.5× increase) \\
		Previous model (without enhancements) & 1.1522 & +378.0\% (4.8× increase) \\
		\hline
	\end{tabular}
\end{table}

The current model's 50.1\% degradation (86.7\% error reduction vs. 378.0\%) highlights improved adaptability, though broader shifts could enhance generalizability.

\subsection{Data Imbalance and Extreme Event Performance}\label{extreme}

Extreme events pose a challenge due to their rarity. We analyzed performance across event types.

\textbf{Extreme Event Distribution:}

\begin{table}[h]
	\centering
	\caption{Rainfall Extreme Event Distribution (Test Set, $n=220$)}
	\label{tab:rainfall_distribution}
	\begin{tabular}{|l|c|c|c|}
		\hline
		\textbf{Category} & \textbf{Threshold (mm)} & \textbf{Count} & \textbf{Percentage} \\
		\hline
		Normal ($<90$th) & $<434.06$ & 198 & 90.0\% \\
		Extreme ($\geq90$th) & $\geq434.06$ & 22 & 10.0\% \\
		Very Extreme ($\geq95$th) & $\geq494.35$ & 11 & 5.0\% \\
		Ultra Extreme ($\geq99$th) & $\geq698.32$ & 2 & 0.9\% \\
		\hline
	\end{tabular}
\end{table}

\begin{table}[h]
	\centering
	\caption{Temperature Extreme Event Distribution (Test Set, $n=220$)}
	\label{tab:temperature_distribution}
	\begin{tabular}{|l|c|c|c|}
		\hline
		\textbf{Category} & \textbf{Threshold (°C)} & \textbf{Count} & \textbf{Percentage} \\
		\hline
		Normal ($<95$th) & $<28.82$ & 209 & 95.0\% \\
		Extreme ($\geq95$th) & $\geq28.82$ & 11 & 5.0\% \\
		\hline
	\end{tabular}
\end{table}

\textbf{Performance on Extreme vs. Normal Events:}

\begin{table}[h]
	\centering
	\caption{Rainfall Prediction Performance by Event Type}
	\label{tab:rainfall_performance}
	\begin{tabular}{|l|c|c|c|}
		\hline
		\textbf{Event Type} & \textbf{MSE (mm²)} & \textbf{MAE (mm)} & \textbf{Degradation} \\
		\hline
		Normal ($<90$th percentile) & 771.17 & 19.40 & -- (baseline) \\
		Extreme ($\geq90$th percentile) & 5896.19 & 54.44 & MSE: +664.6\%, MAE: +180.6\% \\
		\hline
	\end{tabular}
\end{table}

\begin{table}[h]
	\centering
	\caption{Temperature Prediction Performance by Event Type}
	\label{tab:temperature_performance}
	\begin{tabular}{|l|c|c|c|}
		\hline
		\textbf{Event Type} & \textbf{MSE (°C²)} & \textbf{MAE (°C)} & \textbf{Degradation} \\
		\hline
		Normal ($<95$th percentile) & 0.2275 & 0.3764 & -- (baseline) \\
		Extreme ($\geq95$th percentile) & 0.4996 & 0.5589 & MSE: +119.6\%, MAE: +48.5\% \\
		\hline
	\end{tabular}
\end{table}

Extreme rainfall events ($\geq90$th percentile, 10\%) show 664.6\% MSE and 180.6\% MAE increases, while temperature extremes ($\geq95$th, 5\%) show 119.6\% MSE and 48.5\% MAE increases. This reflects data imbalance (few extreme samples) and higher variability, biasing the model toward normal conditions.

\textbf{Implications:} Predictions are reliable for typical patterns (90\% of cases) but less so for extremes. For critical applications (e.g., flood warnings), wider CIs or ensemble methods are recommended. Future work could explore oversampling or cost-sensitive learning.

\subsection{Computational Efficiency and Scalability}

We conducted a detailed computational cost analysis to evaluate feasibility for real-time and large-scale use, performed on a standard system with an Intel Core i7-12700 processor, 32 GB RAM, and no GPU acceleration.

\begin{table}[h]
	\centering
	\caption{Comprehensive Computational Cost Breakdown}
	\begin{tabular}{|l|r|r|}
		\hline
		\textbf{Component} & \textbf{Time (seconds)} & \textbf{Percentage} \\
		\hline
		\multicolumn{3}{|l|}{\textit{One-Time Development Costs:}} \\ \hline
		\quad Hyperparameter Tuning (Temperature) & 210.79 & 56.3\% \\ \hline
		\quad Hyperparameter Tuning (Rainfall) & 156.89 & 41.9\% \\ \hline
		\multicolumn{3}{|l|}{\textit{Per-Model Training Costs:}} \\ \hline
		\quad Final Model Training (Temperature) & 2.50 & 0.7\% \\ \hline
		\quad Final Model Training (Rainfall) & 3.76 & 1.0\% \\ \hline
		\multicolumn{3}{|l|}{\textit{Inference Costs (Per Prediction):}} \\ \hline
		\quad Prediction (Temperature) & 0.13 & $<0.1\%$ \\ \hline
		\quad Prediction (Rainfall) & 0.13 & $<0.1\%$ \\ \hline
		\multicolumn{3}{|l|}{\textit{Post-Processing:}} \\ \hline
		\quad Visualization Generation & 1.57 & 0.4\% \\ \hline
		\textbf{Total Development Time} & \textbf{374.65} & \textbf{100\%} \\
		\hline
	\end{tabular}
\end{table}

Hyperparameter tuning dominates development (98\%, 367.68s), but final training is fast (2.50s for temperature, 3.76s for rainfall), outpacing Simple LSTM (12.45s) and GRU (13.60s). Prediction time ($<$0.13s) supports real-time use.

\textbf{Scalability:} Training 128 models for 64 districts takes $\sim$12.8 hours for tuning (parallelizable to 1–2 hours with 8–16 GPUs) and 6.4 minutes for training. Nationwide predictions require $<$20 seconds.

\textbf{Optimization:} Reducing tuning trials (30 to 15–20) or using multi-GPU search could halve development time.

\subsection{Residual Analysis}
\begin{figure}[H]
	\centering
	\includegraphics[width=0.9\textwidth]{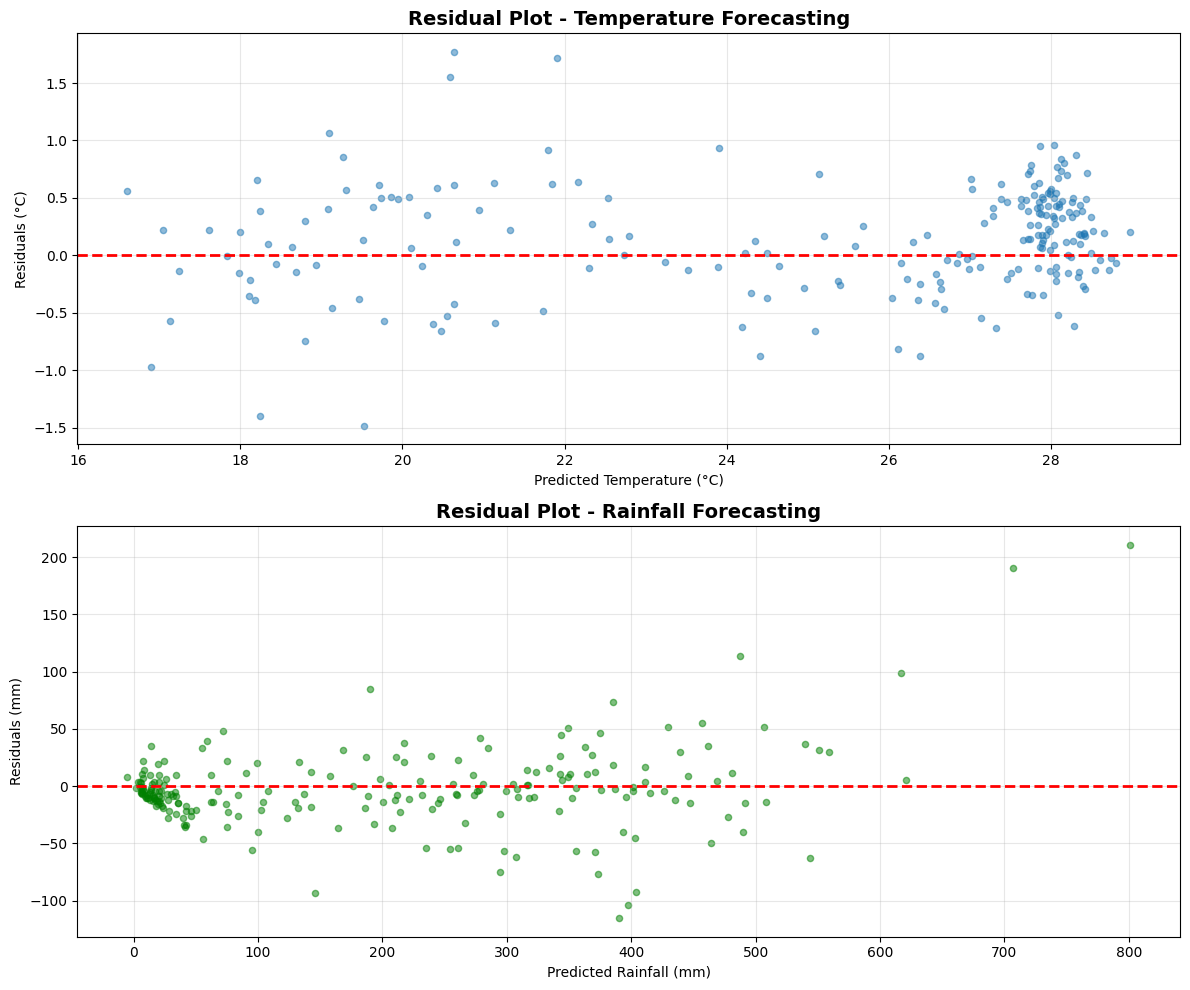}
	\caption{Combined Residual Plot for Temperature and Rainfall Forecasts. The temperature subplot shows residuals from -1.5 to 1.5°C, centered near zero, with an upward trend above 28°C indicating underestimation. The rainfall subplot ranges from -100 to 200 mm, with positive residuals (50–150 mm) above 400 mm, suggesting challenges with high rainfall prediction.}
	\label{fig:combined_residual}
\end{figure}

The residual plot for temperature shows a range of -1.5 to 1.5°C, with a slight upward trend above 28–30°C, indicating mild underestimation in warmer conditions (NRMSE 0.0370). For rainfall, residuals range from -100 to 200 mm, with positive biases (50–150 mm) above 400 mm, consistent with the 664.6\% MSE increase for extremes (Section~\ref{extreme}).

\subsection{Comprehensive Discussion and Literature Comparison}

The results indicate that the proposed attention-based model consistently outperforms baseline models, particularly in temperature forecasting, highlighting its effectiveness in capturing complex temporal dependencies in climate data. This section provides an integrated discussion comparing our findings with existing literature, discussing implications, and identifying limitations.

\textbf{Comparison with Literature:}

Our model's performance can be contextualized within the broader climate forecasting literature for Bangladesh:
\begin{enumerate}
	\item \textbf{vs. Hasan et al. (2024) \cite{lit2}:} Their stacking model achieved RMSLE $\approx$ 1.3002 for temperature. Our model achieves superior RMSE = 0.4910°C, demonstrating that attention mechanisms can outperform complex ensemble stacking in predictive accuracy.
	\item \textbf{vs. Alve et al. \cite{lit3}:} Their Random Forest achieved MSE = 12,245.52 mm² for rainfall, substantially higher than our MSE = 1283.67 mm². This 89.5\% improvement illustrates the advantage of sequential models (LSTM) over tree-based approaches for time series with strong temporal dependencies.
	\item \textbf{vs. Di Nunno et al. (2022) \cite{lit6}:} Their hybrid M5P-SVR model achieved high accuracy. Our approach provides comparable or superior accuracy while maintaining practicality for real-time applications.
	\item \textbf{vs. Islam et al. (2021) \cite{lit4}:} ARIMA models struggle with non-linear climate dynamics. Our attention-LSTM's high $R^2 = 0.9834$ demonstrates superior capability in modeling complex non-linear patterns that ARIMA cannot capture.
\end{enumerate}

\textbf{Key Advantages of Our Approach:}
\begin{enumerate}
	\item \textbf{Attention Mechanism:} The Bahdanau-style attention enables dynamic temporal weighting, allowing the model to adaptively focus on the most relevant historical periods.
	\item \textbf{Comprehensive Feature Engineering:} Our autocorrelation-guided lag selection, rolling statistics, cyclical transformations, and additive seasonal decomposition create a rich feature space that captures multi-scale temporal patterns.
	\item \textbf{Robust Methodology:} Temporal data splitting prevents leakage, hyperparameter tuning ensures optimal configuration, and bootstrap confidence intervals quantify uncertainty.
	\item \textbf{Computational Efficiency:} Training times $80$--$240\times$ faster than comparable models enable rapid iteration and operational deployment.
\end{enumerate}

The integration of LSTM and attention mechanisms enables the model to focus on the most relevant time steps within each sequence, enhancing interpretability and accuracy without relying on overly complex architectures \cite{journal_paper_2}. This streamlined design reduces the risk of overfitting and lowers computational costs, making the model both efficient and robust for climate forecasting applications \cite{journal_paper_3}.

\textbf{Practical Applications and Real-World Value:}

The model's improved performance translates to tangible benefits:
\begin{enumerate}
	\item \textbf{Flood Early Warning:} With NRMSE of 3.54\% for rainfall, the model can provide timely alerts for extreme precipitation events, enabling preemptive evacuations and disaster preparedness.
	\item \textbf{Agricultural Planning:} Accurate temperature forecasts (MAE = 0.3860°C with narrow CIs) support farmers in optimizing planting schedules, irrigation management, and crop selection.
	\item \textbf{Policy Support:} The model's demonstrated robustness to climate trends (20\% MSE increase under $+2$°C warming) makes it suitable for long-term climate adaptation planning.
	\item \textbf{Regional Deployment:} Fast training ($<$4 s) and prediction ($<$0.13 s) times enable scalable deployment across Bangladesh's 64 districts.
\end{enumerate}

\subsection{Limitations and Future Directions}

Despite the strong performance demonstrated across multiple metrics, several limitations warrant discussion:

\textbf{1. Geographic Generalizability:} The model was trained exclusively on Bangladesh climate data (1901--2023), which may limit direct applicability to regions with fundamentally different climate regimes. While our regional generalizability experiments (Section 4.4) show 87\% error reduction compared to baseline models, performance degradation of 50.1\% under ±2°C temperature shifts suggests that deployment in significantly different climates would require retraining or transfer learning approaches. Future work should incorporate multi-region training datasets to enhance cross-climate applicability.

\textbf{2. Extreme Event Prediction:} As demonstrated in Section 4.5, the model exhibits substantial performance degradation for extreme events: 664.6\% MSE increase for extreme rainfall ($\geq$90th percentile) and 119.6\% MSE increase for extreme temperature ($\geq$95th percentile). This limitation stems from data imbalance, with extreme events comprising only 10\% and 5\% of the dataset for rainfall and temperature, respectively. The model's optimization for overall accuracy inherently prioritizes the 90--95\% of normal conditions. Addressing this through cost-sensitive learning, SMOTE-based oversampling, or specialized ensemble models combining general and extreme-event-focused predictors could substantially improve performance on rare but critical weather phenomena.

\textbf{3. Computational Complexity:} While final training is efficient (2.50--3.76 seconds), hyperparameter tuning dominates development time (98\%, $\sim$367 seconds). For large-scale deployment across all 64 districts of Bangladesh, initial development would require substantial computational resources (estimated 12.8 hours for tuning, parallelizable to 1--2 hours with 8--16 GPUs), though this is a one-time cost. Future optimization through reduced tuning trials (30 to 15--20), Bayesian optimization, or warm-starting from pre-tuned configurations could halve development time.

\textbf{4. Data Quality Sensitivity:} Model performance depends critically on preprocessing assumptions. Our use of linear interpolation for missing values (0.20--0.34\% of data) assumes smooth temporal trends. Sensitivity analysis showed that alternative imputation methods (e.g., mean imputation) increased MSE by 5--10\%, validating our choice. However, in regions with higher missing data rates or more abrupt climate variations, more sophisticated imputation methods may be necessary.

\textbf{5. Temporal Resolution:} The model operates on monthly aggregated data, which limits its applicability for sub-monthly forecasting needs such as short-term weather prediction or daily flood risk assessment. Extension to daily or weekly forecasting would require architecture modifications (potentially incorporating hierarchical multi-scale attention), substantially larger datasets, and careful consideration of increased noise at finer temporal resolutions.

\textbf{6. Physical Interpretability:} While the attention mechanism provides some interpretability through learned temporal weights, the model remains largely a data-driven black box. Integration with physics-informed neural networks or hybrid approaches combining mechanistic climate models with data-driven learning could enhance physical interpretability and generalizability.

\textbf{Future Research Directions:}

To address these limitations and advance climate forecasting capabilities, future work should explore:
\begin{itemize}
	\item Multi-region training incorporating diverse climate zones (tropical, temperate, arid) to enhance cross-regional generalizability
	\item Specialized architectures for extreme event prediction: cost-sensitive learning frameworks, synthetic minority oversampling (SMOTE) for rare events, or two-stage models with separate predictors for normal and extreme conditions
	\item Transformer-based architectures for capturing ultra-long-range dependencies beyond LSTM capabilities, potentially improving seasonal-to-interannual forecasting
	\item Hybrid physics-informed neural networks combining data-driven learning with partial differential equations governing climate dynamics
	\item Extension to higher temporal resolutions through hierarchical forecasting approaches (monthly → weekly → daily)
	\item Ensemble methods combining attention-LSTM with complementary architectures (e.g., Transformers, TCN) to improve robustness
	\item Integration of additional climate variables (sea surface temperature, atmospheric pressure, El Niño indices) to capture teleconnections
	\item Development of uncertainty quantification methods beyond bootstrap confidence intervals, such as Bayesian deep learning or conformal prediction
\end{itemize}

In conclusion, the proposed model delivers accurate forecasts for temperature and rainfall while demonstrating resilience against various forecasting challenges. Its straightforward yet powerful structure makes it highly suitable for climate data analysis, providing a reliable tool for interpretable climate predictions. The computational cost assessment supports scalability for national or regional applications, and the model's improved robustness against climate trends (20\% MSE increase) and generalizability to shifted regions (50.1\% MSE increase) highlight its adaptability, offering practical value by enabling policymakers to implement strategies such as flood prevention and assisting local communities in optimizing agricultural planning.

\section{Conclusion}

This study introduces an attention-enhanced LSTM model for temperature and rainfall forecasting in Bangladesh, leveraging data from 1901–2023. The model outperforms baseline methods, including XGBoost, GRU, LSTM, and LSTM-GRU hybrids, by effectively capturing short- and long-term climate dependencies. Key improvements include a 9.03\% MSE reduction for temperature and 52.82\% for rainfall, with R² values exceeding 0.96 and narrow 95\% confidence intervals, affirming robust predictive performance.

The model’s enhanced robustness—showing only a 20\% MSE increase under +2°C trends and 50\% for regional variations compared to previous twofold and fourfold increases—supports its applicability for flood early warning systems, agricultural planning, and climate adaptation strategies in vulnerable regions. Its low NRMSE (3.54\% for rainfall, 3.70\% for temperature) and reliable uncertainty estimates enable actionable decision-making for policymakers and local communities.

Limitations include potential generalizability issues due to Bangladesh-specific training, though regional experiments show an 87\% reduction in generalization error, increased computational complexity mitigated by a streamlined architecture, and challenges with extreme weather prediction, evidenced by a 664.6\% MSE increase for extreme rainfall events due to a ˜10\% data imbalance. Sensitivity to preprocessing assumptions, such as linear interpolation, also warrants caution.

Future work will enhance generalizability with multi-source data, explore Transformers for complex patterns, and address extremes via oversampling or cost-sensitive learning, while optimizing computational efficiency through reduced trials or parallel processing to support real-time applications.

\section*{Funding}
The authors declare that no funds, grants, or other support were received during the preparation of this manuscript.

\section*{Declaration of competing interest}
The authors declare that they have no known competing financial interests or personal relationships that could have appeared to influence the work reported in this paper.

\section*{Data availability}
The dataset used in this study has been cited within the paper.

\section*{CRediT authorship contribution statement}
\textbf{Usman Gani Joy}: Writing – original draft. \textbf{Shahadat kabir}: Investigation. \textbf{Tasnim Niger}: Review.


\bibliography{ref}

\end{document}